\documentclass[runningheads,a4paper]{llncs}

\usepackage{amssymb}
\usepackage{graphicx}
\usepackage{tabularx}
\usepackage{array}
\usepackage{float}
\usepackage{placeins}

\usepackage{url}
\urldef{\mailsa}\path|{alfred.hofmann, ursula.barth, ingrid.haas, frank.holzwarth,|
\urldef{\mailsb}\path|anna.kramer, leonie.kunz, christine.reiss, nicole.sator,|
\urldef{\mailsc}\path|erika.siebert-cole, peter.strasser, lncs}@springer.com|
\newcommand{\keywords}[1]{\par\addvspace\baselineskip
\noindent\keywordname\enspace\ignorespaces#1}

\usepackage[colorlinks=true,
linkcolor=blue,
citecolor=blue,
urlcolor=blue]{hyperref}

\begin{document}

\mainmatter  

\title{The Path Ahead for Agentic AI: Challenges and Opportunities}

\titlerunning{The Path Ahead for Agentic AI}

\author{
Nadia Sibai\inst{1} \and
Yara Ahmed\inst{1} \and
Serry Sibaee\inst{1} \and
Sawsan AlHalawani\inst{1} \and
Adel Ammar\inst{1}\thanks{Corresponding author: aammar@psu.edu.sa} \and
Wadii Boulila\inst{1}
}

\authorrunning{Sibai et al.}

\toctitle{The Path Ahead for Agentic AI: Challenges and Opportunities}
\tocauthor{Sibai et al.}

\institute{
Robotics and Internet-of-Things (RIOTU) Lab, Prince Sultan University, Riyadh, Saudi Arabia\\
\email{\{221410294, 221410920, ssibaee, shalawani, aammar, wboulila\}@psu.edu.sa}
}

\maketitle

\begin{abstract}
The evolution of Large Language Models (LLMs) from passive text generators to autonomous, goal-driven systems represents a fundamental shift in artificial intelligence. This chapter examines the emergence of agentic AI systems that integrate planning, memory, tool use, and iterative reasoning to operate autonomously in complex environments. We trace the architectural progression from statistical models to transformer-based systems, identifying capabilities that enable agentic behavior: long-range reasoning, contextual awareness, and adaptive decision-making. The chapter provides three contributions: (1) a synthesis of how LLM capabilities extend toward agency through reasoning-action-reflection loops; (2) an integrative framework describing core components perception, memory, planning, and tool execution that bridge LLMs with autonomous behavior; (3) a critical assessment of applications and persistent challenges in safety, alignment, reliability, and sustainability. Unlike existing surveys, we focus on the architectural transition from language understanding to autonomous action, emphasizing the technical gaps that must be resolved before deployment. We identify critical research priorities, including verifiable planning, scalable multi-agent coordination, persistent memory architectures, and governance frameworks. Responsible advancement requires simultaneous progress in technical robustness, interpretability, and ethical safeguards to realize potential while mitigating risks of misalignment and unintended consequences.

\end{abstract}


\section{Introduction}

Language has long been central to artificial intelligence (AI), shaping how machines interpret, generate, and interact through natural language. 
Early natural language processing (NLP) relied on handcrafted rules and basic statistical models, which required explicit programming for each task. 
The emergence of Large Language Models (LLMs), which are AI systems trained on massive text corpora, marked a significant shift in artificial intelligence. 
These models are designed upon transformer-based architectures that utilize attention mechanisms to process and relate information in sequences, enabling strong generalization, instruction-following, and emergent reasoning capabilities. 
As a result, LLMs have evolved into flexible cognitive engines capable of performing a wide range of tasks, including text summarization, code generation, dialogue, and complex problem-solving.

As LLMs have grown in scale and capability, they have become integrated into real-world systems such as ChatGPT, Gemini, Claude, and LLaMA, driving widespread adoption in multidisciplinary fields such as education, industry, and research. However, this rapid progress also exposes critical limitations, including but not limited to: high computational demands, opaque decision processes, and challenges related to bias, misinformation, and accountability \cite{devlin2019bert}.
These gaps highlight the need for AI systems that go beyond text generation towards more structured, transparent and controllable forms of intelligence. This need is addressed in recent agentic AI frameworks that integrate structured planning, tool use, modular decision pipelines, and human-in-the-loop control \cite{yao2023react,AbouAli2025,Plaat2025}. This, in turn, improves transparency, controllability, and accountability in real-world deployments.

This chapter examines the shift from passive LLMs to agentic AI systems. These systems can plan, act, use tools, review outcomes, and use feedback loops. Agentic AI goes beyond single-turn responses to autonomous, goal-driven behavior supported by memory, reasoning, and interaction with the environment. Understanding this shift requires technical grounding and a critical examination of the current architectures, capabilities, and limitations.

Accordingly, the chapter is organized as follows. In Section~2, we trace the historical progression of LLMs, highlighting architectural milestones relevant to the agency. Next, Section~3 introduces the core principles of agentic AI. Section~4 then explains the integrative architectures that bridge LLM reasoning with planning, memory, and tool use. Following this, Section~5 surveys applications in various domains. Finally, Section~6 discusses challenges and outlines directions for future research.

The chapter makes three contributions, summarized as follows:
\begin{enumerate}
    \item A brief synthesis of how LLMs move naturally toward agentic behavior;
    \item A clear framework detailing components and feedback loops in agentic AI;
    \item A critical review of key applications, along with open technical, ethical, and research challenges.
\end{enumerate}

Together, these contributions aim to provide both a foundational primer and a forward-looking perspective on the path ahead for agentic AI.

\section{History of LLMs}

The development of large language models (LLMs) reflects decades of progress in modeling language, scaling architectures, and refining training strategies. 
Each major paradigm shift, including statistical models, neural networks, recurrent architectures, transformers, and large-scale pre-training, introduced new capabilities. These capabilites now underpin agentic behavior, such as long-range reasoning, contextual awareness, and adaptive decision-making. \textbf{Table~\ref{tab:timeline}} summarizes these milestones.

\subsection{Statistical Language Models (1990s)}
Statistical n-gram models \cite{rosenfeld2000slm} provided an early foundational approach for probabilistic sequence prediction. However, they were fundamentally limited by sparse data, short context windows, and rigid probability estimation. 
Although they lacked semantic reasoning, these models established the core principle of next-word prediction that later evolved into richer forms of planning and action selection in agentic systems.

\subsection{Neural Language Models (2000s)}
Neural probabilistic language models \cite{bengio2003neural} introduced distributed word embeddings and continuous representations, enabling generalization beyond observed text. These models captured semantic similarity and contextual patterns more effectively than n-grams. Though constrained by fixed context windows, this representational shift laid the groundwork for more expressive mechanisms for reasoning processes which are essential for agentic behavior.

\subsection{Recurrent Networks and Embeddings (2010s)}
Recurrent Neural Networks (RNNs) and Long Short-Term Memory (LSTM) models \cite{mikolov2010rnn} enhanced the effective context length through learned internal memory, while Word2Vec \cite{mikolov2013word2vec} produced scalable and reusable embeddings. 
Although these architectures faced training inefficiencies and struggled to capture very long-range dependencies \cite{jozefowicz2016exploring}, they  introduced foundational mechanisms for maintaining temporal continuity and task grounding features, which  later became essential for agents that require multi-step planning and persistent contextual awareness.

\subsection{Transformer Models and Pre-training (late 2010s)}
The transformer architecture \cite{vaswani2017attention} replaced recurrence with \textit{self-attention}, enabling parallel computation, global context integration, and scalable depth. 
Innovations such as positional encodings, multi-head attention, and early forms of sparse attention \cite{child2019generatinglongsequencessparse} supported long-range reasoning. 
Pre-trained models such as BERT \cite{devlin2019bert} and GPT-2 \cite{radford2019gpt2} demonstrated that large corpora combined with fine-tuning could yield versatile language systems. These architectural breakthroughs form the cognitive backbone of agentic AI: global context tracking, structured reasoning traces, and the ability to invoke multi-step thought sequences.

\subsection{The Era of Large-Scale LLMs (2020s–present)}
The 2020s marked a shift toward large-scale models guided by scaling laws, extensive training pipelines, and advanced alignment techniques. Models such as GPT-3 \cite{brown2020gpt3}, PaLM \cite{chowdhery2022palm}, and LLaMA \cite{touvron2023llama} leveraged billions of parameters, Mixture-of-Experts (MoE) routing \cite{mu2025comprehensivesurveymixtureofexpertsalgorithms}, and optimized data pipelines to achieve strong few-shot generalization and emergent reasoning. Reinforcement Learning from Human Feedback (RLHF) \cite{lambert2025reinforcementlearninghumanfeedback} and instruction tuning enabled goal-directed behavior and safer interaction, while multimodal models such as GPT-4 \cite{achiam2023gpt4} integrated vision and language capabilities. These models introduced core agentic capabilities such as tool use, planning, and self-reflection which transitions LLMs from passive generators to systems capable of autonomous, context-aware action.

\begin{table}[ht]
\centering
\caption{Four decades of progress in language modeling and their relevance to agentic AI.}
\label{tab:timeline}
\renewcommand{\arraystretch}{1.45}
\begin{tabularx}{\linewidth}{l|X|X|X}
\hline
\textbf{Decade} & \textbf{Key Advances} & \textbf{Representative Models} & \textbf{Relevance to agentic AI} \\
\hline
1990s &
Statistical LMs; smoothing/back-off &
n-gram SLMs \cite{rosenfeld2000slm} &
Established sequence prediction as a core task, enabling later planning and decision-making.
\\
2000s &
Neural LMs; distributed representations &
Neural LM \cite{bengio2003neural}, RNN LMs \cite{mikolov2010rnn} &
Introduced semantic and contextual representations, supporting richer reasoning for agent behavior.
\\
2010s &
Embeddings; Transformers; pre-training &
word2vec \cite{mikolov2013word2vec}, Transformer \cite{vaswani2017attention}, BERT \cite{devlin2019bert}, GPT-2 \cite{radford2019gpt2} &
Transformers enabled global context, structured reasoning, and scalable pre-training foundations of modern agentic systems.
\\
2020s &
Scaled LLMs; instruction tuning; multimodal models &
GPT-3 \cite{brown2020gpt3}, PaLM \cite{chowdhery2022palm}, LLaMA \cite{touvron2023llama}, GPT-4 \cite{achiam2023gpt4} &
Large-scale models exhibit emergent reasoning, tool use, and autonomy, which are essential for planning and adaptive agent behavior.
\\
\hline
\end{tabularx}
\end{table}


\section{Agentic AI: Concepts, Examples, and Architectures}

Agentic AI refers to  systems capable of autonomous decision-making, tool use, planning, and adaptive behavior to achieve specific goals \cite{Abumalloh2025}. 
Unlike traditional LLMs, which generate one-shot text responses, agentic AI operates through iterative perception–reasoning–action loops. This enables agents to decompose complex tasks, interact with external environments, and refine their actions based on feedback \cite{AbouAli2025}. 
These capabilities such as long-term planning, contextual memory, and tool invocation, enable agents to function as collaborative problem solvers rather than passive text generators. By decomposing tasks into explicit reasoning, action, and reflection steps, agentic architectures make intermediate decisions more observable and auditable. This  partially addresses concerns about opaque decision-making and accountability compared to monolithic (i.e., single ) LLM inference system \cite{yao2023react,AbouAli2025}.

\subsection{Evolution from Classical to LLM-based AI Agents}

Classical symbolic AI, also known as rule-based AI, represents knowledge explicitly and performs deterministic reasoning to achieve goals. Examples include Belief–Desire–Intention (BDI) models and sense–plan–act (SPA) pipelines, which rely on structured rules and explicit world models. While these systems provided structured and rule-based reasoning, they struggled to operate well in open-ended environments. 
Large language models (LLMs) introduced a new form of generative AI, which is capable of producing coherent text but largely passive in their behavior. 
 These models employ pattern-based learning and statistical reasoning to achieve natural language understanding and text generation. However, they usually work in a  single-turn interaction following a simple prompt-generate-respond cycle, without iterative refinements.
On the other hand, modern agentic systems built on large language models (LLMs) leverage \emph{stochastic} and prompt-driven reasoning techniques such as chain-of-thought, reflective refinement, and tool-calling strategies \cite{Plaat2025}. This allows them to generate flexible and context-aware actions. This shift from fixed rules to generative reasoning enables agents to operate autonomously in open-ended environments which  transforms LLMs from passive producers into goal-directed and adaptive agents. Table \ref{tab:ai_evolution_comprehensive} summarizes  a comparison for the evolution of the different AI paradigms. 

Agents operating within an agentic AI system are generally classified into three functional categories based on their roles and capabilities \cite{Plaat2025}, which are:
\textbf{(1) Reasoning agents}   which perform internal cognition such as reflection, goal decomposition, and memory-based planning.
\textbf{(2) Action agents}   that interface with tools, APIs, or robotic systems to perform concrete tasks.
\textbf{(3) Multi-agent or interactive systems}   which coordinate multiple agents through communication, negotiation, or role specialization.
Hybrid systems increasingly combine these capabilities, integrating reasoning, memory, and tool-based execution.

\begin{table}[!htbp]
\centering
\caption{Comprehensive Comparison: Evolution from classical symbolic AI to large language models (LLMs) and modern agentic AI, highlighting the transition from rule-based decision-making to generative language understanding that supports iterative reasoning and action loops.}
\label{tab:ai_evolution_comprehensive}
\small
\begin{tabular}{|p{2.1cm}|p{3.3cm}|p{3.3cm}|p{3.3cm}|}
\hline
 \textbf{Dimension} & \textbf{Classical Symbolic AI} & \textbf{Large Language Models} & \textbf{Modern agentic AI} \\
\hline
\hline

\multicolumn{4}{|c|}{\rule[-1.3ex]{0pt}{4ex} \textbf{Core Characteristics}} \\
\hline

\textbf{Reasoning} & Deterministic & Pattern-Based & Stochastic \\
\hline

\textbf{Knowledge} & Explicit Rules & Learned Patterns & Prompt-Driven \\
\hline

\textbf{Agency} & Reactive & Passive & Goal-Directed \\
\hline

\textbf{Adaptability} & Limited & Moderate & High \\
\hline

\textbf{Environment} & Structured & Single-Turn & Open-Ended \\
\hline
\hline

\multicolumn{4}{|c|}{\rule[-1.3ex]{0pt}{4ex} \textbf{Key Features}} \\
\hline

& 
• Deterministic reasoning \newline
• Explicit rules \newline
• Logical inference \newline
• Structured problem-solving & 
• Prompt-based generative text \newline
• Pattern-based learning \newline
• Language understanding \newline
• Few-shot learning & 
• Goal-directed behavior \newline
• Chain-of-thought reasoning \newline
• Tool integration \newline
• Reflective refinement \\
\hline
\hline

\multicolumn{4}{|c|}{\rule[-1.3ex]{0pt}{4ex} \textbf{Operation and Performance}} \\
\hline

\textbf{Operation Flow} & 
Perceive $\rightarrow$ Decide $\rightarrow$ Execute & 
Prompt $\rightarrow$ Generate & 
Observe $\rightarrow$ Reason $\rightarrow$ Act $\rightarrow$ Reflect $\circlearrowleft$ \\
\hline

\textbf{Limitations} & 
Limited adaptability; fragile under uncertainty & 
Passive; no goal pursuit; single-turn responses & 
Stochastic outputs; probabilistic reasoning; requires careful prompting \\
\hline

\textbf{Advantages} & 
Transparent; predictable; interpretable & 
Flexible; broad knowledge; natural language interface & 
Autonomous; adaptive; goal-directed \\
\hline

\end{tabular}
\end{table}


\subsection{Single-Agent vs. Multi-Agent Agentic AI}

Agentic AI may involve a single autonomous agent or a coordinated multi-agent system \cite{AbouAli2025,Sapkota2025}.
In the single-agent architecture, one agent is responsible for executing the entire task pipeline end-to-end, including perception, reasoning, and action. However, multi-agent systems decompose the task into multiple specialized roles each is handled by distinct agents, which improves modularity, scalability, and robustness for complex or multi-stage problems. This distinction is summarized in Table~\ref{tab:agentic-contrast}.











\begin{table}[!htbp]
\centering
\caption{Agent (single autonomous LLM) vs. agentic AI system (multi-agent orchestration).}
\label{tab:agentic-contrast}
\renewcommand{\arraystretch}{1.1}
\begin{tabularx}{\linewidth}{@{}lXX@{}}
\hline
\textbf{Aspect} & \textbf{Single Agent} & \textbf{Agentic AI System (Multi-agent)} \\
\hline
Scope & Completes a task end-to-end & Decomposes tasks across specialized agents \\
Coordination & None; self-contained loop & Role assignment, scheduling, negotiation \\
Memory & Local/episodic store & Shared/long-term memory; vector DB or KB \\
Tool use & Calls tools/APIs directly & Tooling + inter-agent tool delegation \\
Failure modes & Single-point failure & Coordination errors; cascading failures \\
Evaluation & Task success and cost & Team metrics: throughput, reliability, auditability \\
Examples & ReAct-style single agent \cite{yao2023react} & AutoGen/Crew-style teams \cite{Wu2023,AbouAli2025} \\
\hline
\end{tabularx}
\end{table}
\FloatBarrier

\section{Bridging LLMs and Agentic AI}

Large Language Models (LLMs) form the cognitive core of modern agentic AI systems. While the LLM provides high-level reasoning, planning, and decision-making, an external agent framework supplies the complementary components needed for autonomy: perception, memory, action execution, and environmental interaction. Together, these elements form a closed-loop control architecture in which the LLM does not merely generate text but continuously plans, acts, and adapts based on feedback \cite{yao2023react,Schick2023}.

\subsection{Core Components of an Agentic Architecture} \label{sec:components}

\begin{figure}[h]
\centering
\includegraphics[width=1.05\linewidth]{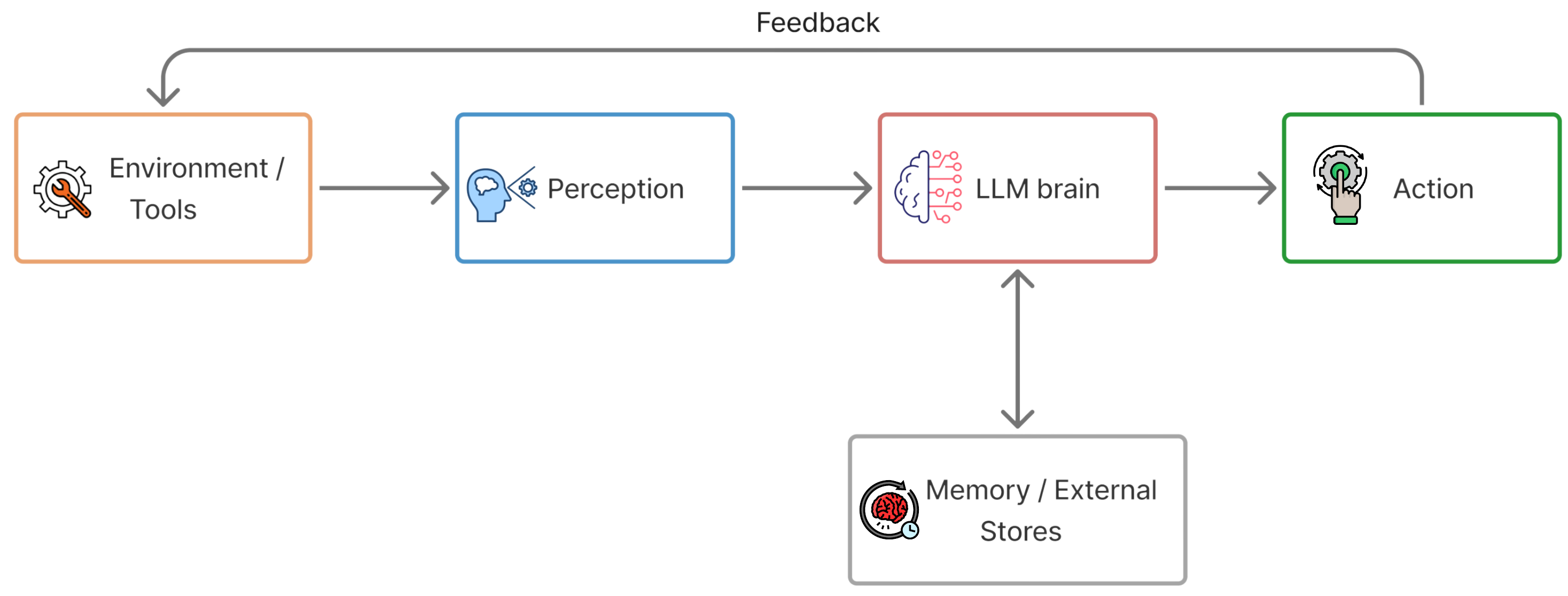}
\caption{The core components of an agentic AI system that operate within a continuous feedback loop.
The Environment/Tools provide execution capabilities, Perception handles raw observations, the LLM Brain performs reasoning and planning while  Memory enables persistence,  and Action executes plans. Arrows indicate the flow of information and control through this cycle.}
\label{fig:Architecture_Diagram}
\end{figure}

Figure~\ref{fig:Architecture_Diagram} illustrates the interaction among key components.
Each module plays a distinct role in enabling agency:

\begin{itemize}
\item \textbf{Environment / Tools:}
external systems which the agent interacts with, such as: APIs, search engines, calculators, robots, databases, software tools, or simulated environments. These components provide grounded information that influences the agent's decision-making process and affects the agent’s outcomes.

\item \textbf{Perception:}  
the mechanism that converts raw observations (tool outputs, sensor data, retrieved documents) into structured input for the LLM. This may include text parsing, multimodal interpretation, or result summarization.

\item \textbf{LLM Brain (Reasoning and Planning):}  
the central cognitive engine responsible for chain-of-thought reasoning, goal decomposition, tool selection, and action planning. Here, the LLM determines \emph{what to do next} based on context and prior steps.

\item \textbf{Memory / External Stores:}  
episodic or long-term memory enables persistence across steps or sessions.  
Examples include vector databases, scratchpads, and domain-specific knowledge bases. Recent work on Retrieval-Augmented Generation (RAG) demonstrates that hyperparameter optimization in vector stores, chunking strategies, and re-ranking mechanisms significantly impacts both retrieval quality and system efficiency, with implications for memory-augmented agentic systems \cite{ammar2025rag}.  
Memory supports long-horizon tasks, identity consistency, and iterative refinement.

\item \textbf{Action:}  
execution of plans through API calls, tool invocation, code execution, robotic control, or other operations. Actions feed results back into the environment, completing the feedback loop.

\end{itemize}

These components of the agentic AI architecture operate in a cyclical feedback loop that  enables continuous learning and adaptation through repeated perception-reasoning-action cycles.

\subsection{The Reason–Act–Reflect Loop (Single-Agent Systems)} \label{Sec:ReAct}

Most agentic systems follow a recurrent \textit{reason–act–reflect} pattern:
\begin{enumerate}
\item \textbf{Reason:} The LLM interprets the current state, decomposes tasks, and decides the next action.
\item \textbf{Act:} An external tool, API, or environment module executes the chosen action.
\item \textbf{Reflect:} The LLM reviews the outcome, updates memory, corrects errors, and adjusts its plan.
\end{enumerate}

Frameworks like \textbf{ReAct} \cite{yao2023react} and \textbf{Toolformer} \cite{Schick2023} implement this loop by explicitly interleaving reasoning traces with tool calls. This transforms a static LLM into an adaptive, interactive agent.

\subsection{Conversational Multi-Agent Systems } \label{sec:multi-agent}

This approach features the involvement of  multiple agents that communicate and coordinate with each other to accomplish certain tasks. The framework enables the use of customizable agents that can seamlessly integrate LLMs, human inputs and tools in various configurations. Using different agent interaction patterns through both natural language and programmatic control enables agents to coordinat and share information effectively. AutoGen \cite{Wu2023} framework is an example that follows this multi-agents paradigm.

\subsection{Agent Frameworks Enabling LLM Integration}

Recent frameworks such as \textbf{LangChain} and \textbf{AutoGen} operationalize these ideas. Both frameworks are open-source tools that are widely used in the context of agentic AI and LLM applications. \textbf{LangChain} focuses on single-agent control, tool integration, and memory while \textbf{AutoGen}  focuses on multi-agent coordination, communication, and collaborative task execution.  They enable agentic AI  by providing:

\begin{itemize}
\item standardized tool interfaces for APIs, search engines, and code execution;
\item memory modules for long-term retrieval and episodic context;
\item orchestrators for multi-agent collaboration and role assignment;
\item guardrails for safe execution and workflow monitoring.
\end{itemize}

The LLM issues natural-language instructions, while the framework manages reliable execution \cite{Wu2023,AbouAli2025}. This synergy enables practical, domain-specific agentic AI systems in education, research, automation, and robotics.





\section{Concrete Examples of agentic Systems}

In order to provide a clearer understanding of agentic AI, this section presents  concrete examples that illustrate the paradigm in practice.  These examples highlight both single-agent and multi-agent architectures, demonstrating how planning, tool usage, memory management, and coordination are implemented in real-world scenarios. Through such  examples, the abstract concepts of agentic AI become more tangible, helping to bridge the gap between theoretical frameworks and practical applications.

\subsection{Single-Agent Example} 
\begin{figure}[!htbp]
    \centering
    \includegraphics[width=\linewidth]{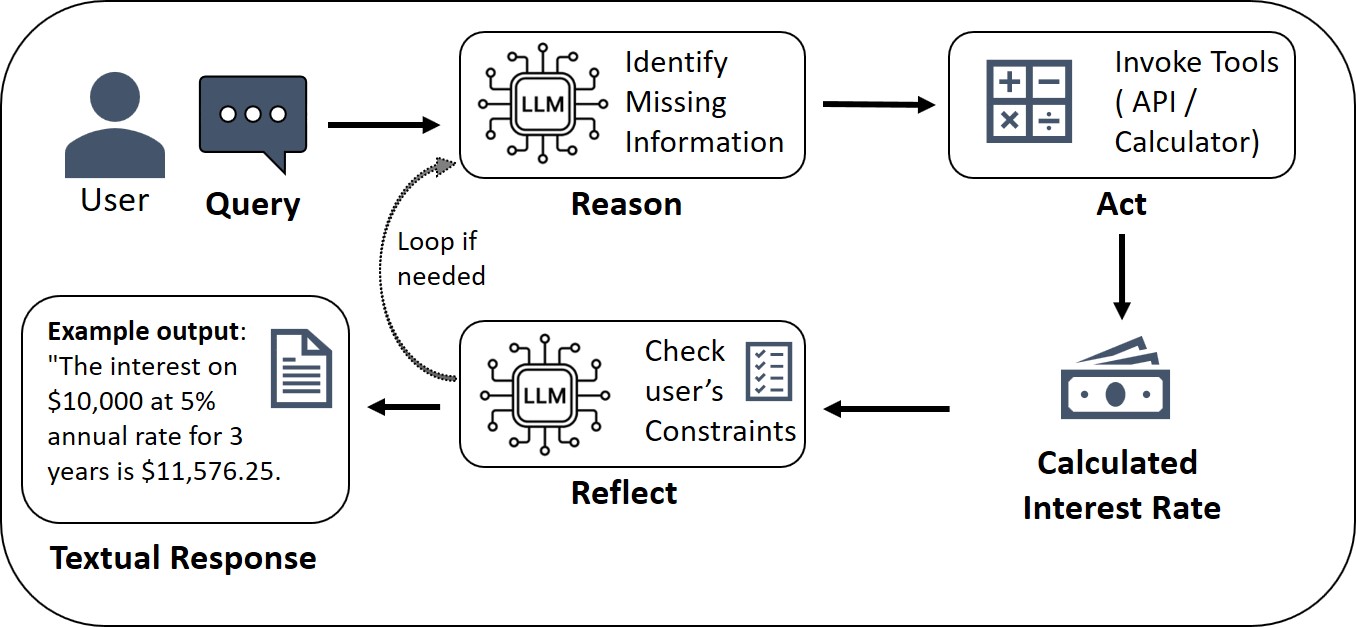}
    \caption{Single-agent iterative ReAct architecture for financial query processing. The LLM iteratively reasons about missing information, acts by invoking tools, and reflects on results before generating the final response.}
    \label{fig:ReAct}
\end{figure}

This example illustrates the use of the (ReAct-style) \cite{yao2023react} that uses the Reason-Act-Reflect cycle explained in section \ref{Sec:ReAct} to process financial queries.
A ReAct agent solving a financial query may involve:
\begin{enumerate}
\item Reason: Identify missing information needed.
\item Act: Use a calculator or API to compute interest.
\item Reflect: Verify whether the result meets the user’s constraints.
\end{enumerate}
Here, a single LLM drives planning and tool usage within a closed loop \cite{yao2023react}. Figure \ref{fig:ReAct} shows an iterative ReAct pattern for financial query processing. The Reason-Act-Reflect cycle repeats until all user constraints are satisfied before generating the final textual response.

\subsection{Multi-Agent Example }
\begin{figure}[H]
    \centering
    \includegraphics[width=\linewidth]{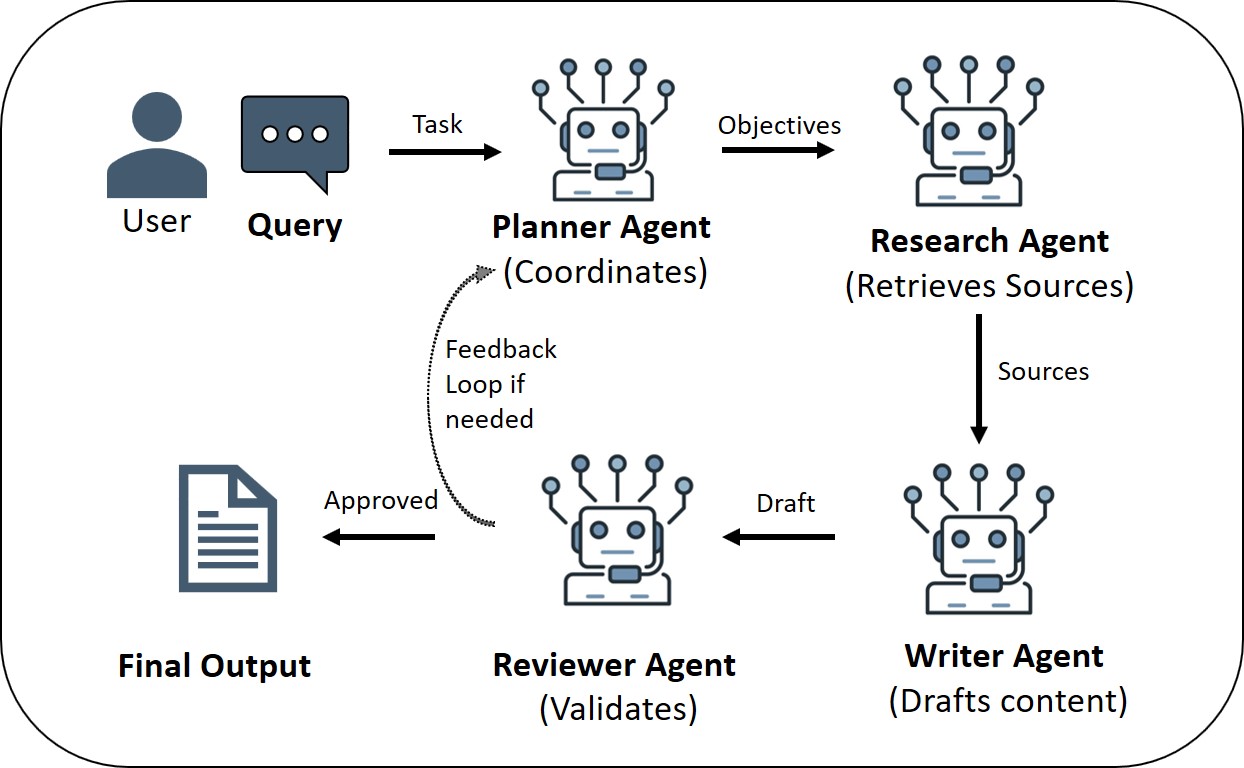}
    \caption{A multi-agent system that employs four specialized agents in sequence: a Planner that coordinates objectives, a Research agent that retrieves sources, a Writer that drafts content, and a Reviewer that validates output quality. When validation fails, a feedback loop enables iterative refinement by returning control to the Planner.}
    \label{fig:MultiAgent}
\end{figure}

This example follows  the (AutoGen-style) \cite{Wu2023} explained in section \ref{fig:MultiAgent}.
A multi-agent system performing a small research task may involve multiple agents, each having a distinct role, including:
\begin{itemize}
\item A \emph{Planner agent} to outline objectives,
\item A \emph{Research agent} to retrieve and summarize sources,
\item A \emph{Writer agent} to synthesize content, and
\item A \emph{Reviewer agent} to check consistency and correctness.
\end{itemize}
These agents communicate iteratively, exchanging summaries and feedback until the final output is produced \cite{song2025adaptiveinconversationteambuilding}.
Such workflows illustrate how agentic AI extends beyond a single system to coordinated teams. 
Figure \ref{fig:MultiAgent} visualizes a multi-agent AutoGen-style workflow for research tasks. This architecture is an example of coordinated team-based agentic AI, where specialized capabilities are distributed across multiple agents rather than concentrated in a single system.

\subsection{End-to-End Research Workflow}

In order to relate the single-agent paradigm with the components of an agentic system, the following simplified example illustrates a ReAct-style workflow using the architecture shown in Figure \ref{fig:Architecture_Diagram}:

\begin{enumerate}
\item The user inputs a query to ask: “Summarize the latest findings on lithium-ion battery degradation.”
\item \textbf{Reason:} The LLM identifies missing information and decides to search the web.
\item \textbf{Act:} Using a search API, the agent retrieves recent papers.
\item \textbf{Perception:} Retrieved text is cleaned and summarized before being fed back into the LLM.
\item \textbf{Memory:} Key findings are stored in a vector database for long-term reference.
\item \textbf{Reflect:} The LLM verifies whether more data is needed.  If it is insufficient, the agent loops back to Step 2 (Reason) in order to refine the query and get additional sources. Otherwise, it proceeds to generate the final output.
\item \textbf{Final Output:} The agent synthesizes insights into a structured technical summary.
\end{enumerate}

This example highlights how planning, tool use, and memory interlock within a coherent agentic workflow. However, while it is presented sequentially for clarity, perception is responsible on processing both the initial query and subsequent tool outputs. Memory is accessed throughout reasoning steps.

Important agentic examples include AutoGPT \cite{autogpt2023}, BabyAGI \cite{babyagi2023}, Voyager for embodied skill acquisition \cite{Wang2023}, and Toolformer for self-supervised API use \cite{Schick2023}. In human-robot interaction, ROSGPT \cite{koubaa2025rosgpt} demonstrates how LLMs can translate unstructured natural language into structured robotic commands through prompt engineering and ontology-based interpretation, bridging conversational AI with physical action execution. These systems reveal the core properties of agentic AI: planning, tool-use, persistent memory, and iterative reasoning.

\section{Challenges and Future Directions}
\label{sec:challenges}

Despite rapid progress, agentic AI faces substantial technical, ethical, and operational challenges rooted in its ability to pursue goals and perform real-world actions. Ensuring reliability, safety, and sustainability requires advances in architecture, governance, and evaluation \cite{zeng2025applicationdriven,gridach2025agentic}.

\subsection{Safety, Alignment, and Control}

Ensuring that autonomous agents behave consistently with human intentions is a central challenge. Unlike static LLMs, agentic systems can initiate actions, place orders, modify code, or trigger workflows, making misalignment potentially consequential \cite{lynch2025agentic}.
This risk is evident in several real-world scenarios, for example: a financial agent might misinterpret a liquidity event and automatically liquidate assets, or a customer service agent might issue refunds incorrectly after misreading logs \cite{nacar-etal-2025-towards}.

The reliable deployment of agentic AI systems requires robust evaluation frameworks that extend beyond surface-level metrics. Recent work on Arabic language model evaluation reveals significant gaps in existing benchmarks, particularly in linguistic accuracy, cultural alignment, and methodological rigor, with leading models achieving only 30\% accuracy on culturally grounded reasoning tasks \cite{sibaee2025arabic}. 
These findings highlight the broader challenge of developing comprehensive evaluation methodologies capable of assessing agentic systems across diverse linguistic and cultural contexts \cite{sibaee-etal-2024-asos-ksaa}.

In order to address the risks associated with agentic AI systems, it is important to implement mitigation strategies  that assure  safety, alignment and control. The main purpose of utilizing these strategies is to ensure transparent decision-making processes, reduce potential harmful behavior and maintain human oversight where necessary. The following are key approaches commonly used as mitigation strategies:

\begin{itemize}
\item \textbf{Controllable autonomy:} restricting agent permissions through role-based, time-bounded, and context-aware execution constraints.
\item \textbf{Structured guardrails:} integrating policy-enforecement tools, safety layers and reversible actions to prevent unsafe behavior.
\item \textbf{Auditability:} enabling traceability through mandatory chain-of-thought logs, action justification, and rollback capabilities.
\item \textbf{Human-in-the-loop checkpoints:} requiring supervised approvals for high-impact or safety-critical actions.
\end{itemize}

Scaling these safeguards to open-ended, multi-objective environments remains an open problem.

\subsection{Reliability and Robustness}

The reliability of agentic systems depends on stable planning, accurate tool use, and consistent multi-step reasoning. In practice, agents face challenges including:

\begin{itemize}
\item \textbf{Long action chains:} multi-step workflows amplify small errors at different stages, making system behavior more difficult to predict and  debugging process more challenging \cite{zhang2025agenticcontextengineeringevolving}.
\item \textbf{Non-deterministic behavior:} stochastic decoding, probabilistic reasoning and variable responses  from external APIs can result in different outputs for similar inputs. This reduces reproducibility and increases  uncertainty. 
\item \textbf{Opaque components:} using closed-source models hinders verification, transparency and external auditing.
\end{itemize}

Research on verifiable reasoning, uncertainty calibration, and hybrid symbolic–neural systems aims to mitigate these challenges without sacrificing adaptability \cite{gridach2025agentic}. Empirical studies comparing LLM performance with human experts in complex programming tasks reveal that, although LLMs excel at certain pattern-matching activities, they score significantly lower than humans in multi-step problem-solving challenges. This highlights the persistent limitations in the reliability of agentic reasoning  \cite{koubaa2023humans}.

\subsection{Memory and Long-Term Consistency}

Persistent memory enables long-horizon tasks but introduces risks of drift, hallucinated recall, privacy leakage, and compounding biases. Current agents struggle to maintain consistent identities, plans, or task states over extended interactions.


Several mitigation strategies could be utilized to avoid the risks introduced by persistent memory in agentic AI such as: 
\begin{itemize}
\item \textbf{Hierarchical memory architectures:} separating short-term working memory, episodic memory, and long-term knowledge stores, can reduce interference across extended interactions.
\item \textbf{Episodic recall mechanisms:} retrieving memories based on context and particular tasks, interactions, or time-frames rather than relying on mixed long-term memory  maintains contextual accuracy, consistency and  reduces hallucination. 
\item \textbf{Controlled forgetting and correction:} this includes relevance filtering, confidence thresholds, and decay functions which can prevent the accumulation of outdated or low-quality information.
\item \textbf{Selective retention and anonymization mechanisms:} handling sensitive or biased information can be managed through data sanitization (e.g., anonymization or abstraction) or removed entirely to prevent privacy leakage and bias propagation.
\end{itemize}

\subsection{Ethical, Legal, and Societal Implications}

Agentic AI raises questions of accountability, transparency, and responsible autonomy. When semi-autonomous agents take actions that are economically or socially impactful, traditional responsibility boundaries become unclear.
Case examples include agents making unauthorized trades, generating discriminatory recommendations, or bypassing internal approval processes.

Key mitigation strategies include:
\begin{itemize}
\item \textbf{Enforceable explainability requirements:} ensuring agents provide clear and traceable justifications for their decisions.
\item \textbf{Standardized audit logs and oversight protocols:} enabling consistent monitoring and post-hoc analysis of agent actions.
\item \textbf{Human override mechanisms:} allowing operators to intervene or halt agent behavior when risks happen.
\item \textbf{Regulatory frameworks for autonomous systems:} defining legal accountability and compliance obligations for agent-driven decisions.
\end{itemize}

These safeguards are essential to maintain trust and prevent systemic harms.

\subsection{Computational and Environmental Costs}

Agentic AI architectures, including long interaction loops, frequent tool calls, and continuous context expansion, significantly increase computational requirements beyond standard LLM inference.
Training and deploying such systems raise important sustainability concerns, motivating research into model compression, adaptive inference, and hardware-efficient execution \cite{refai2025peeringinsideblackbox}.
Therefore, sustainable engineering must become a core design principle for future agentic systems.

\subsection{Future Research Agenda}

To address the  challenges discussed above, several promising research directions are emerging such as the following:

\begin{enumerate}
\item \textbf{Reliable Planning and Tool Usage:} developing robust action modeling, verifiable execution, and recovery mechanisms to ensure agents can behave safely and reliably. 
\item \textbf{Scalable Interpretability:} creating real-time introspection tools, transparent action traces, and interpretable policies to understand autonomous agents and analyse their behavior especially in complex environments. 
\item \textbf{Continuous and Structured Memory:} designing long-term episodic memory, adaptive retrieval, and models that preserve consistency across extended interactions over weeks or months.
\item \textbf{Multi-Agent Coordination Frameworks:} establishing protocols for communication, negotiation, division of labor, and conflict resolution to facilitate agents' collaboration in team-based tasks. 
\item \textbf{Efficient and Sustainable Inference:} developing energy-aware agent architectures, dynamic model selection, and low-overhead tool orchestration to reduce computational cost and improve environmental sustainability
\item \textbf{Governance and Auditing Infrastructure:} implementing standardized safety tests, alignment benchmarks, permission systems, and regulatory guardrails to acheive accountable and trustworthy autonomous behavior.
\end{enumerate}

\section{Conclusion}



This chapter examined the transformative shift from passive Large Language Models to agentic AI systems capable of autonomous planning, tool use, and adaptive decision-making. By tracing architectural evolution from statistical n-grams through transformer-based pre-training to contemporary agentic frameworks, we demonstrated how fundamental capabilities, such as global context integration, emergent reasoning, and iterative refinement, naturally extend toward goal-directed behavior. The integrative architecture presented, centered on perception, reasoning, action, and memory feedback loops, provides a conceptual foundation for understanding how LLMs transition from text generation to autonomous operation.

Our analysis reveals that while current systems demonstrate impressive capabilities in task decomposition and multi-step reasoning, fundamental challenges still exist. Safety and alignment concerns intensify when agents initiate real-world actions beyond text generation. Reliability issues can accumulate across long action chains where small errors propagate and amplify at each step. Memory systems struggle to maintain consistency across extended interactions, increasing the risk of drift and hallucinations. These technical limitations intersect with ethical questions about accountability and transparency when decision boundaries blur between human operators and autonomous systems.

\textbf{Forward-Looking Research Priorities.} We identify three critical directions for advancing agentic AI:

\begin{enumerate}
\item \textbf{Hybrid Symbolic-Neural Architectures:} combining symbolic planning with LLM-based reasoning could enable verifiable action traces, bounded behavior guarantees, and interpretable decisions while preserving adaptability.

\item \textbf{Hierarchical Multi-Agent Coordination:} protocols for inter-agent communication, role negotiation, conflict resolution, and dynamic task decomposition are essential as systems scale beyond single-agent workflows.

\item \textbf{Sustainable and Adaptive Inference:} energy-aware architectures, dynamic model selection, and efficient context management through sparse attention and retrieval-augmented generation will determine responsible scalability.

\end{enumerate}

This chapter provides a structured synthesis of how LLM capabilities enable agentic behavior, an integrative architectural framework, and a critical assessment of current limitations, offering both a technical primer and a research roadmap. The path ahead requires parallel advances in governance, auditing infrastructure, and alignment methodologies to ensure autonomous systems remain controllable and aligned with human values.

The long-term vision involves systems functioning as reliable and transparent collaborators capable of sustained reasoning and ethical decision-making. Achieving this demands continued research into robust planning, interpretable policies, persistent memory, and efficient execution, alongside regulatory frameworks that establish clear accountability. Only through this integrated approach, balancing innovation with ethical safeguards and sustainability, can agentic AI augment human capabilities while minimizing risks of misalignment and societal harm.

\bibliographystyle{unsrt}
\bibliography{references}

\end{document}